\documentclass[10pt,journal,compsoc]{IEEEtran}
\ifCLASSOPTIONcompsoc
  \usepackage[nocompress]{cite}
\else
  \usepackage{cite}
\fi

\ifCLASSINFOpdf
  \usepackage[pdftex]{graphicx}
\else
\fi

\usepackage{amsmath}

\usepackage{url}

\usepackage{amsfonts} 
\usepackage{pgfplots}
\usepackage{subcaption}
\usepackage{makecell}
\usepackage{multirow}

\usepackage{lipsum}

\newcommand\blfootnote[1]{%
	\begingroup
	\renewcommand\thefootnote{}\footnote{#1}%
	\addtocounter{footnote}{-1}%
	\endgroup
} 

\begin{document}

\title{VertiBayes: Learning Bayesian network parameters from vertically partitioned data with missing values}
\author{Florian~van Daalen,~\IEEEmembership{Graduate Student Member,~IEEE,}
        Lianne~Ippel,
        Andre~Dekker,
        and~ Inigo~Bermejo% <-this % stops a space
\IEEEcompsocitemizethanks{\IEEEcompsocthanksitem  F. van Daalen, I. Bermejo, and A. Dekker are with 	the Department of Radiation Oncology (MAASTRO) GROW School for Oncology and Reproduction Maastricht University Medical Centre+ Maastricht the Netherlands\protect\\
\IEEEcompsocthanksitem L. Ippel is with Statistics Netherlands Heerlen the Netherlands. }% <-this % stops a space
}


\IEEEtitleabstractindextext{%
\begin{abstract}
		Federated learning makes it possible to train a machine learning model on decentralized data. Bayesian networks are widely used probabilistic graphical models. While some research has been published on the federated learning of Bayesian networks, publications on Bayesian networks in a vertically partitioned data setting are limited, with important omissions, such as handling missing data. We propose a novel method called VertiBayes to train Bayesian networks (structure and parameters) on vertically partitioned data, which can handle missing values as well as an arbitrary number of parties. For structure learning we adapted the K2 algorithm with a privacy-preserving scalar product protocol. For parameter learning, we use a two-step approach: first, we learn an intermediate model using maximum likelihood, treating missing values as a special value, then we train a model on synthetic data generated by the intermediate model using the EM algorithm. The privacy guarantees of VertiBayes are equivalent to those provided by the privacy preserving scalar product protocol used. We experimentally show VertiBayes produces models comparable to those learnt using traditional algorithms. Finally, we propose two alternative approaches to estimate the performance of the model using vertically partitioned data and we show in experiments that these give accurate estimates.
\end{abstract}

\begin{IEEEkeywords}
Federated Learning, Bayesian network, privacy preserving, vertically partitioned data, missing values, parameter learning, structure learning
\end{IEEEkeywords}}

%\IEEEpeerreviewmaketitle
\maketitle

% is not selected <OR> if conference mode is selected - because compsoc
% conference papers position the abstract like regular (non-compsoc)
\IEEEdisplaynontitleabstractindextext

% \fi

\ifCLASSOPTIONcompsoc
\IEEEraisesectionheading{\section{Introduction}\label{sec:introduction}}
\else
\section{Introduction}
\label{sec:introduction}
\fi


\blfootnote{\emph{The views expressed in this paper are those of the authors and do not necessarily reflect the policy of Statistics Netherlands.}}

\IEEEPARstart{F}{ederated} learning is a field that recently rose to prominence due to the increased focus on data-hungry techniques, privacy concerns and protection of the data\cite{li_review_2020}\cite{kairouz_advances_2021}. Using federated learning, it is possible to train a machine learning model without needing to collect the data centrally \cite{li_review_2020}. Since it rose to prominence, various techniques for training a model on centrally collected data have been adapted to be used on data that is either horizontally or vertically partitioned\cite{kairouz_advances_2021}. Data is said to be horizontally partitioned if multiple parties collect the same variables though from different individuals, e.g., two hospitals who want to build a model to predict heart failure. It is said to be vertically partitioned when multiple parties collect different variables about the same individuals, for example, data from a hospital and from a health insurance company where both parties have unique variables about the same patients.
 
A type of model that can benefit from federated learning are Bayesian networks. Bayesian networks are probabilistic graphical models that have been widely used in artificial intelligence \cite{pearl_probabilistic_2014}. They are popular because they can be built, verified, or improved, by combining data with existing expert knowledge. For example, medical doctors can manually create the network structure, ensuring it models already known dependencies correctly, while the conditional probability distributions are estimated from data. Thanks to its graphical representation and probabilistic reasoning, it is also a relatively intuitive model for non-technical personnel. This makes it very useful in scenarios where non-technical personnel needs to make decisions based on the model, for example when used as a tool to inform healthcare policies.

While research has been published on federated learning of Bayesian networks \cite{yang_privacy-preserving_2006} \cite{wright_privacy-preserving_2004} \cite{zhiqiang_yang_improved_2005} \cite{ng_towards_2021}, publications on Bayesian networks trained on vertically partitioned data (also referred to as heterogeneous data in the literature) are limited. For example, to the best of our knowledge there are no published approaches to train a Bayesian network from vertically partitioned data with missing values. Furthermore, the published work on vertically partitioned data, has focused on scenarios involving only two parties. In this article, we propose a novel method called VertiBayes to train Bayesian networks on vertically partitioned data, which can handle missing values as well as an arbitrary number of parties. We will then discuss the strengths and limitations of our proposal.

\section{Methods}
In this section, we will start by shortly explaining how a Bayesian network is generally trained in a central setting. After this we will discuss how the various methods used to build and validate a Bayesian network can be adapted to a vertically partitioned federated setting.
		
\subsection{Training a Bayesian network}
Training a Bayesian network consists of two phases: structure learning and parameter learning. The first phase, structure learning, consists of determining the structure of the graph (i.e., the set of links between variables) and can be done either manually, using expert knowledge, or automatically, using algorithms such as K2\cite{cooper_bayesian_1992}. The second phase is the so-called parameter learning. In this phase, the conditional probability distributions (CPDs) for each node in the network are determined. Throughout this paper, we will focus on CPDs in the form of conditional probability tables (CPTs) as these are the most common form of CPD. In the next subsections, we will discuss how this is done in a centrally trained scenario. After which we will discuss how these methods need to be adapted for the federated scenario.

\subsubsection{Structure learning}
The structure of a Bayesian network can be either determined manually or learnt using an algorithm. Here, we focus on the latter, since the former does not involve data analysis. One of the most popular structure learning algorithms is K2, which performs a heuristic search for a viable structure by scoring potential parent nodes for a given node and step-wise adding the highest scoring parent\cite{cooper_bayesian_1992}. The scoring function used in K2 is described in equation \ref{K2 scoring function} below. 

\begin{equation}
	\label{K2 scoring function}
	f(i,\pi_{i})=\prod_{j=1}^{q_{i}}\frac{(r_{i}-1)!}{(N_{ij} + r_{i}-1)!}\prod_{k=1}^{r_{i}}a_{ijk}!\\
\end{equation}
where $pi_{i}$ is the set of parents of node $x_{i}$. $q_{i}$ is the number of possible instantiations of the parents of $x_{i}$ present in the data. $r_{i}$ is the number of possible values the attribute $x_{i}$ can take. $a_{ijk}$ is the number of cases in the dataset where $x_{i}$ has it's $k^{th}$ value and the parents are initiated with their $j^{th}$ combination. $N_{ij} = \sum_{k=1}^{r_{i}} a_{ijk}$, is the number of instances where the parents of $x_{i}$ are initiated with their $j^{th}$ combination.	

It is important to note that the resulting structure depends on the order in which nodes are introduced into the K2 algorithm. As such, it is possible to construct different structures for the same data.

\subsubsection{Parameter learning}
There are two relevant scenarios to consider when performing parameter learning:  with and without missing data.  When there is no missing data, CPDs can be learned using the maximum likelihood \cite{koller_probabilistic_2009}. To calculate the maximum likelihood for an attribute $X$ with a set of parents $Y$ one simply has to calculate:
$P(X=x_i| Y_i=y_i) = N_(x_i,y_i )/N_(y_i )$ , 
where $N_(x_i,y_i )$ is the number of records where $X=x_i$ and  $Y_i=y_i$ and $N_(y_i )$ is the number of records where $Y_i=y_i$. 
In the presence of missing data, the maximum likelihood for training a Bayesian network is commonly estimated using algorithms such as Expectation Maximization (EM)\cite{dempster_maximum_1977}\cite{lauritzen_em_1995}. The EM algorithm consists of the following two steps repeated iteratively until convergence is reached:
\begin{enumerate}
	\item Estimate the likelihood of your data using your current estimates of the probabilities.
	\item Update your estimates.
\end{enumerate}
It is important to note that since this algorithm is a hill climbing-type algorithm, it can get stuck in local optima. Therefore, it is good practice to run the algorithm several times with different random initializations and use the best result\cite{koller_probabilistic_2009}.

\subsection{Federated learning in a vertically split setting}
\label{federated learning}
Training a Bayesian network in a vertically partitioned federated setting introduces additional hurdles and concerns. First, we will discuss how to perform structure learning. In the second subsection, we will discuss parameter learning.

\subsubsection{Structure learning}
As mentioned previously, structure learning can be done using the K2 algorithm. In this subsection, we will discuss how to adapt the K2 algorithm to a vertically partitioned scenario. To solve this equation, the following information needs to be collected:
\begin{enumerate}
	\item The number of possible values for the attribute $X$.
	\item The number of instances that fulfill $X=x_{i}$ and $Y =y_{j}$, where $X$ is the child attribute, $x_i$ is a given value for $X$, $Y$ is the set of parent attributes, and $y_j$  is a given set of assigned values to $Y$. This needs to be calculated for every possible set $j$.
\end{enumerate}
The number of possible values of attribute $X$ can be calculated trivially without revealing any important information to an external party in a vertically split federated setting as all relevant information is available locally at one party.

To calculate the number of instances that fulfill $X=x_{i}$ and $Y =y_{i}$ (the number of instances for each possible value of $X$ for each possible configuration of $X$’s parents) one has to calculate the number of instances that fulfill certain conditions across different datasets. Fortunately, this problem can be solved using a privacy preserving scalar product protocol and calculating the scalar product of vectors, one for each site, where each individual is represented as $1$ or $0$ depending on whether they fulfill the local conditions (in this case whether the child and parent nodes have the appropriate values). Earlier research has used this approach to calculate the information-gain when training a decision tree\cite{du_building_2002}, which at its root, poses the same problem we face here.

Various variants of the privacy preserving scalar product protocol have been published \cite{du_building_2002} \cite{du_privacy-preserving_2001} \cite{goos_secure_2001} \cite{hutchison_private_2005} \cite{vaidya_privacy_2002}. Most of these focus on $2$-party scenarios but variants do exist for $N$ parties\cite{van_daalen_privacy_2022}. These methods have different advantages and disadvantages, such as different privacy guarantees and risks, different runtime complexities, and different communication cost overheads. Because of this, the preferred method will differ per scenario. A K2 implementation using one of these protocols will have the same privacy guarantees and risks but will pose no additional privacy concerns beyond those posed by the chosen protocol.

\subsubsection{Parameter learning}
During parameter learning, the actual CPDs will be calculated. There are two scenarios that need to be considered: with and without missing values. EM works under the assumption that data is missing at random or missing completely at random.

\paragraph{Without missing values}
As discussed earlier, parameter learning without missing values can be done by calculating the maximum likelihood for various attribute values. This means calculating for each node $i$, $N_{ij}$, the number of samples for each possible configuration of the parents of node $i$ and $N_{ijk}$ the number of samples for each possible configuration of the parents where the value is $k$, for each possible value of the node. $N_{ijk}$ can be calculated by simply summing the various $N_{ij}$ values. For the sake of performance it is advised to do this. These can be calculated using the scalar product protocol as explained earlier when describing the solution for K2. As such, performing parameter learning in a vertically split federated scenario with no missing values is not a problem and can be done without any significant additional privacy risks compared to the central variant beyond the risks involved in the scalar product protocol implementation used.

\paragraph{With missing values} 
In the presence of missing data, it is not possible to use the maximum likelihood and one has to use more sophisticated approaches such as the EM algorithm. As discussed earlier, this algorithm consists of two steps repeated iteratively. The E-step where the likelihood of the training data given the current estimates of the CPDs is calculated, and a M-step where the current estimates are updated.

To estimate the likelihood of the current estimates in the E-step the following equation needs to be solved:
\begin{equation}
	E=\prod_{i=1}^{n}P(d_i)=\prod_{i=1}^{n}\prod_{j=1}^{p}P(x_{ij}|y_{ij})
\end{equation}
where $n$ is the number of samples in the dataset, $p$ is the number of nodes in the network, $P(d_i)$ is the likelihood of the $i-th$ sample, $x_{ij}$ is the value of the $j-th$ node in the $i-th$ sample, and $y_{ij}$ is the set of values of the parents of the $j-th$ node in the $i-th$ sample. The appropriate values $P(x_{ij} |y_{ij} )$ need to be selected from the current estimate of the CPDs based on the attribute values of this particular sample.

However, selecting these in a vertically split federated setting is not trivial without revealing individual level data. In principle, it is possible to use the scalar product protocol to select the appropriate probability from the CPD without revealing the attribute values directly. This can be done by representing the CPD as one of the vectors used in the scalar product protocol, while the other vectors consists of $0$’s and $1$’s representing if this particular cell in the CPD is applicable given the local data each party has. Unfortunately, it should be noted that the selected probability from the CPD may not be revealed, as the probability might be unique in the table, which would allow a party to determine the values of the parents and the node simply by looking up the probability in the CPD. This means that the practical utility of this alternative is limited to scenarios in which the final model is never published.

Revealing the selected probability could be prevented by using fully homomorphic encryption and only decrypting once $P(d_i)$ has been calculated. However, this is impractical due to the high runtime complexity of current fully homomorphic encryption schemes, the runtime complexity of a privacy preserving scalar product protocol, the large number of scalar product protocols which would need to be executed for each $P(d_i)$, and the need to repeat this process for each E-step.

Therefore, we conclude that the EM algorithm cannot be easily applied in a vertically split federated scenario without severe limitations. Instead, we propose the following three-step solution, which we have dubbed VertiBayes.
\begin{enumerate}
	\item Treat “missing” as a valid value and train an intermediate Bayesian network using maximum likelihood on the training data.
	\item Generate synthetic data (including “missing” values) using this intermediate Bayesian network
	\item Train the final model on this synthetic data using the EM algorithm
\end{enumerate}
As discussed earlier, parameter learning in a vertically split federated setting without missing values is possible with the privacy guarantees provided by the privacy preserving scalar product protocol used. Generating synthetic data by using this intermediate model also does not add any additional privacy concerns compared to a centrally trained model, as this is a basic functionality of any Bayesian network. On the contrary, the final model has a reduced risk of data leak because it is trained on synthetic data\cite{berlingerio_privacy_2019}\cite{kairouz_advances_2021}.

\begin{figure}
	\centering
	\includegraphics[scale=0.6]{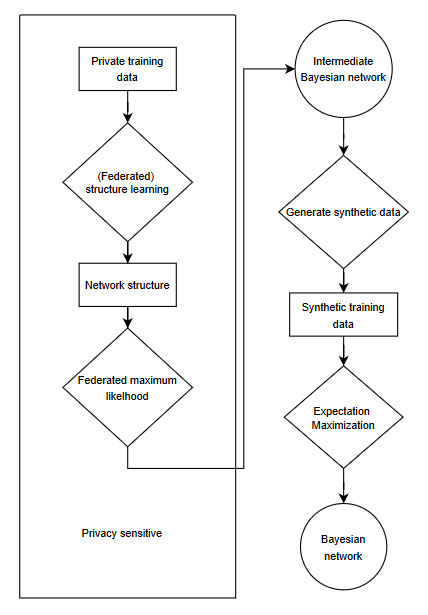}
	\caption{Training process for VertiBayes}
	\label{training}
\end{figure}	

The proposed process, which is illustrated in Figure \ref{training}, allows us to train a Bayesian network in a vertically split federated setting with missing values without any additional privacy concerns compared to a centrally trained model. However, it should be noted that it is possible that a loss of signal may occur due to the three-step approach. In the experiment section, we will test if our proposed method avoids this potential pitfall.

\subsection{Time complexity when training a model in a federated setting}\label{timesection}
A major downside to federated learning is that the time complexity is usually considerably worse compared to the centralized setting. This is unavoidable due to the extra overhead created by communication as well as the increased complexity introduced by the privacy preserving mechanisms. In this subsection, we will discuss the time complexity of VertiBayes.

In our implementation, there are two important factors to take into account. The first is the number of parties $n$. This is a major bottleneck as the $n$-party scalar product protocol implementation we have used scales combinatorically in the number of parties. 

The second important factor is the size of the CPDs that need to be calculated, as each unique probability that needs to be calculated requires a separate $n$-party scalar product protocol to be solved. As such, our implementation scales linearly in the number of probabilities that need to be calculated. The time complexity of various aspects for our implementation can be found in table \ref{time complexity}.

\begin{table*}[h]
	\tiny
	\centering
	\caption{Time complexity}
	\label{time complexity}	
	\begin{tabular}{c|c}
		Number of Scalar product protocols &  \makecell[c]{$O(m)$, \\ where $m$ is the number of unique parent-child value combinations \\ for which a probability needs to be calculate} \\ \hline
		Number of scalar product subprotocols per protocol & \makecell[c]{$\frac{n!}{(x!(n-x)!))}$, for each $x, 2 <= x <=n$, \\ where $n$ is the number of parties involved in the protocol} \\ \hline
		Number of multiplications per subprotocol & \makecell[c]{$O(p*n*(n-1))$, \\ where $p$ is the population size, and $n$ is the number of parties \\ involved in the protocol} \\
	\end{tabular}
\end{table*}

Important to note is that the population size is not a main driver of the runtime. A relatively simple network trained on a small dataset, but with a high number of unique attribute values will have a significantly longer runtime than a more complex network with few unique attribute values. This is because the overall time complexity is dominated by the number of scalar product protocols and subprotocols, which is independent of the population size, but dependent on the number of probabilities that need to be calculated. Finally, it is important to note that there is ample room for parallelization to improve the running time as each scalar product protocol that is needed for VertiBayes is fully independent and can easily be run in parallel.

\subsection{Federated classification and model validation}
The process of using the model to classify new instances in a federated setting is itself a complex problem that depends strongly on the type of model used. In this subsection, we will discuss the methods that are available to classify an individual in a vertically partitioned setting using a Bayesian Network and the implication this has for the validation of the model.

Classification of new samples using a Bayesian Network in a vertically partitioned federated setting suffers from the same issues as the expectation step in the EM algorithm. To classify an instance from vertically partitioned data, we need to select the appropriate probabilities from the CPD. As discussed before, doing this without revealing data to other parties would require a combination of intricate data representation, the scalar product protocol, and fully homomorphic encryption.  While this might be feasible when classifying one instance at a time, it is not practical when classifying a large number of instances in batch or when the Bayesian network is particularly complex. This has major consequences for the validation of a new model in a federated setting. As such, whenever possible the validation should be done using a publicly available dataset which avoids the need for privacy preserving measure during validation. If such a dataset is not available, we propose two different approaches, “Synthetic Cross-fold Validation” (SCV) and “Synthetic Validation Data Generation” (SVDG), to validate the model in a privacy preserving manner.

SCV uses the synthetic data generated by the intermediate Bayesian network as both training and validation data by executing the EM training using k-fold cross validation. However, it is possible that this results in overfitting on the synthetic data and therefore the performance estimate may be biased by the intermediate Bayesian network.

\begin{figure*}
	\centering
	\includegraphics[scale=0.6]{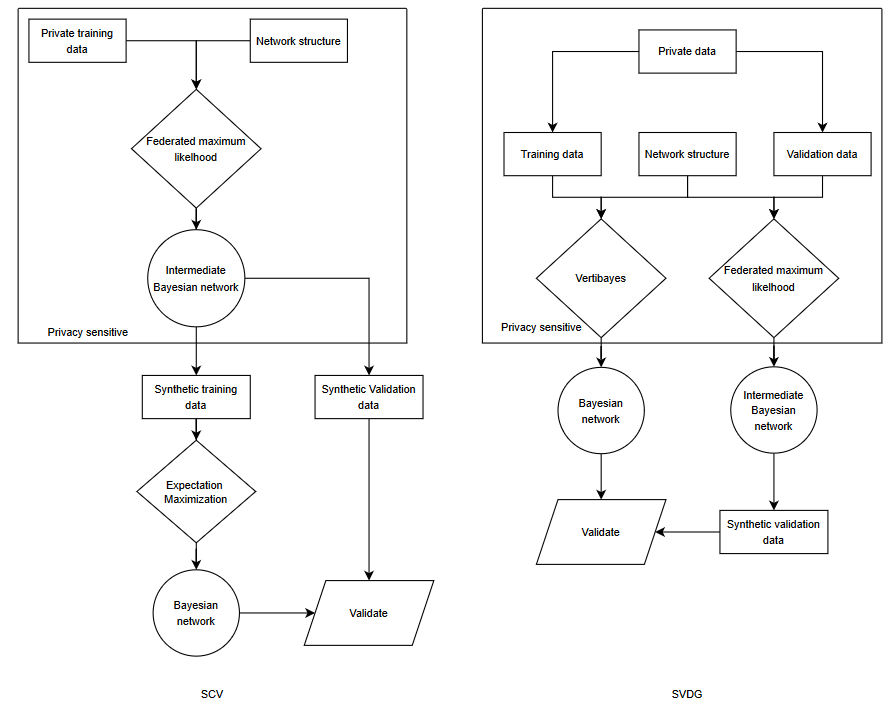}
	\caption{Flow diagrams for proposed validation procedures SCV (left) and SVDG (right)}
	\label{validation}
\end{figure*}

SVDG splits the private dataset into training and validation sets. It will then train a Bayesian network on the training set in a federated manner as normal. On the validation set, it will train a federated network using only the federated maximum likelihood approach. We can then use the Bayesian network trained on the validation set to generate a synthetic validation dataset. This approach reduces the risk of overfitting the previous approach suffered from but may lead to biased estimates if the synthetic validation set is not representative of the original validation set. These approaches avoid leaking real data, but as mentioned, they may not be viable in practice. An illustration of the two new approaches can be seen in Figure \ref{validation}

In the next section, we will perform experiments to validate that VertiBayes results in networks with similar performance as a centrally trained model. We will also verify if the two new approaches to validation give the same results as the validation on the public data, and if there are scenarios where they are inappropriate due to the aforementioned risks.

\subsection{Experimental setup}
In order to validate our proposed approach, we have implementedit in a combination of Java and Python\footnote{
	Our code can be found in the following two git repositories	
	\begin{itemize}\item Main algorithm code: \\ \url{https://github.com/MaastrichtU-CDS/VertiBayes} 
		\item Vantage6 wrapper code: \\ \url{https://github.com/MaastrichtU-CDS/VertiBayes_vantage6}\end{itemize}
} 
using Vantage6\cite{moncada-torres_vantage6_2020} and ran a number of experiments. Vantage6 is an open-source infrastructure for privacy preserving federated learning which utilizes Docker. We compare the performance of our algorithm with a centrally trained Bayesian network using WEKA\cite{frank_data_2016}, a machine learning library written in Java.

\subsubsection{Structure learning}
To validate our federated implementation of the K2 algorithm we ran experiments using the Iris\cite{de_marsico_mobile_2015}, Asia\cite{lauritzen_local_1988}, Alarm\cite{beinlich_alarm_1989}, and Diabetes\cite{smith_using_1988} datasets. As K2 is deterministic and dependent on the order in which the nodes are put into the algorithm we ensured this was the same for both the federatedly and centrally trained model and then compared the resulting structures.

\subsubsection{Parameter learning}
In our experiments regarding parameter learning, we have used the Iris\cite{de_marsico_mobile_2015}, Asia\cite{lauritzen_local_1988}, Alarm\cite{beinlich_alarm_1989}, and Diabetes\cite{smith_using_1988} datasets. In the case of the Iris dataset, we predict the “label” attribute, for Asia we predict “lung”, for Alarm we will be predicting “BP” and for Diabetes we will predict “outcome”. The Iris dataset contains $150$ samples, the Diabetes dataset contains $768$ samples, while the other two datasets contain data of $10.000$ samples. The Asia and Alarm datasets come with a predefined structure. The Iris dataset uses a naïve Bayes structure. The Diabetes dataset also uses a predefined structure.

Both the Iris and Diabetes dataset contain continuous variables. For the sake of a fair comparison between the central and federated models, these were discretized into bins before starting our experiments, where each bin contains at least $10\%$ of the total population as well as a minimum of $10$ individuals. If the last bin cannot be made large enough to fulfill these criteria it is simply added to the previous bin. In the case of a dataset of less than $10$ individuals, the bin will simply contain all possible values. For our experiments the bins were predefined alongside the predefined network structure, but the bins can also be generated on the fly during the training of the federated model using the same discretization strategy. The simplicity of this strategy allows it to be executed without needing additional privacy preserving mechanics. However, it should be noted that this is not the best possible discretizing strategy. For example, a model using the Minimum Description Length method (MDL)\cite{fayyad_multi-interval_1993} for discretization, or utilizing expert knowledge, might outperform this setup.

Slight variations of this discritezation strategy were used in preliminary experiments. However, we will not list the results of those variants here as they produced similar results.

To test the effect of missing values we have done experiments where we randomly set $0\%$, $5\%$, $10\%$ and $30\%$ of the values to missing. The performance of the models is measured by calculating the area under receiver operating characteristics curve (AUC). The centrally trained model is internally validated using $10$-fold cross validation. The federated model is validated using the two different validation approaches described in the last section; it is also validated against a “public” central dataset, which is simply the left-out fold from the original dataset. All of these approaches use $10$-fold cross validation. We also compare the Akaike information criterion (AIC)\cite{akaike_new_1974} values of the network for their original (private) training data. This is done to determine if there is any difference in the CPDs used by the two models.

For all of these experiments the data is randomly partitioned over two parties by dividing the original dataset into two equally large sets of attributes.

\section{Results}

The results of our experiments can be found in this section. First we will discuss the results of the structure learning experiments. After that we will cover the experiments regarding parameter learning.

\subsection{Structure learning}
The federated implementation of the K2 algorithm consistently resulted in the same structure as the centrally trained model for all datasets. 

\subsection{Parameter learning}
Table \ref{experimental results} shows the results of our parameter learning experiments. It lists the AUC for the centrally trained model, as well as the AUC’s for the various (federated) validation schemes. The AUC’s of the federated model are listed in bold if the difference with the central model is larger than $0.05$. The AIC is shown for the centrally trained model, this is compared to the AIC for the federated model trained. The difference between the central and federated AIC is listed in bold if the difference is at least $5\%$. An AIC closer to $0$ is better. A negative percentage in the AIC difference column indicates the federated model performed better.

\begin{table*}[h]
	\tiny
	\centering
	\caption{Results of the experiments.\\
		The experiments are grouped per dataset. The AUC is represented for the centrally trained model, as well as the two federated validation schemes; “Synthetic Cross-fold Validation” (SCV) and “Synthetic Validation Data Generation” (SVDG). AUC values are listed in bold if they differ more than $0.05$ from the central model. An AIC closer to $0$ is better, a negative percentage in the AIC difference collumn indicates the federated model was better. This value is listed in bold if the difference between the federated AIC and the central AIC is at least $5\%$}
	\label{experimental results}	
	\begin{tabular}{c|c|c|c|c|c|c|c|c|c}
		\multicolumn{2}{c}{} \vline & \multicolumn{4}{c}{AUC} \vline & & \multicolumn{3}{c}{AIC} \\ \cline{3-10}
		
		\multicolumn{2}{c}{} \vline & \multicolumn{4}{c}{Training method}\vline &  &\multicolumn{2}{c}{Training Method} \vline &\\ \cline{3-9}
		
		\multicolumn{2}{c}{} \vline & Centrally trained & \multicolumn{3}{c}{Federatedly trained} \vline &  & \multirow{2}{*}{\begin{tabular}{c}Centrally trained\end{tabular}} &\multirow{2}{*}{\begin{tabular}{c}Federatedly trained\end{tabular}} & 	\multirow{2}{*}{\begin{tabular}{c}AIC difference\end{tabular}} \\ \cline{1-6}
		Dataset &  \makecell{Missing data \\ level} & Public validation & Public validation & SCV validation & SVDG validation & & &  & \\ \hline
		\multirow{4}{*}{\begin{tabular}{c}Alarm \\ population size: 10000 \end{tabular}} &0&0,91&0,91&0,91&0,91&&-340571&-318469&\textbf{-6,49}\% \\ 
		&0,05&0,88&0,89&0,88&0,89&&-315856&-313612&-0,71\% \\
		&0,01&0,85&0,86&0,86&0,86&&-340823&-350576&2,86\% \\ 
		&0,3&0,76&0,76&0,76&0,76&&-444297&-444866&0,13\% \\ \hline
		\multirow{4}{*}{\begin{tabular}{c}Asia \\ population size: 10000 \end{tabular}}&0&1,00&1,00&1,00&1,00&&-22555&-22559&0,02\% \\
		&0,05&0,76&0,76&0,76&0,76&&-23430&-23395&-0,15\% \\
		&0,01&0,69&0,7&0,7&0,7&&-24105&-24090&-0,06\% \\
		&0,3&0,58&0,59&0,58&0,59&&-25517&-25613&0,37\% \\ \hline
		\multirow{4}{*}{\begin{tabular}{c}Diabetes \\ population size: 768\end{tabular}}&0&0,8&0,77&\textbf{0,95}&0,79&&-13593&-14407&\textbf{5,99\%} \\
		&0,05&0,79&\textbf{0,74}&\textbf{0,92}&0,76&&-13699&-14556&\textbf{6,25\%} \\
		&0,01&0,75&0,72&\textbf{0,90}&0,73&&-13761&-14408&4,70\% \\
		&0,3&0,61&0,57&\textbf{0,80}&0,6&&-14736&-15136&2,71\%  \\ \hline
		\multirow{4}{*}{\begin{tabular}{c}Iris \\ population size: 150 \end{tabular}}&0&0,99&0,98&1,00&1,00&&-1036&-1022&-1,33\% \\ 
		&0,05&0,97&0,96&0,99&0,97&&-1243&-1176&\textbf{-5,37}\% \\
		&0,01&0,9&0,89&0,95&0,92&&-1491&-1022&\textbf{-31,49}\% \\
		&0,3&0,98&0,99&1,00&0,99&&-1099&-1381&\textbf{25,67\%} \\ \hline
		
	\end{tabular}
\end{table*}

\section{Discussion}
In this paper, we have proposed a novel method to train Bayesian networks in a federated setting using vertically partitioned data with missing values. We have shown that it is possible to perform both structure and parameter learning of Bayesian networks in such a setting with reasonable accuracy. Structure learning can be performed by adapting any of the existing structure learning algorithms to use a secure multiparty computation algorithm. In this study, we used a  protocol within the K2 algorithm, but the same approach could be applied to other score-based algorithms or even constraint-based algorithms such as the PC algorithm\cite{spirtes_algorithm_1991}. Parameter learning requires one of two approaches. When there is no missing data present, the scalar product protocol can be used directly to compute the maximum likelihood, or when missing data is present the three-step solution described in section \ref{federated learning} using the EM algorithm can be applied. We will now discuss the performance of VertiBayes compared to a centrally trained model as well as the limitations in terms of scalability. Lastly, we will discuss the sensitive information that may be leaked by any Bayesian network and the limitations this brings in a federated setting.

\subsection{Model performance and validation}
Our experiments show that the resulting models produced by VertiBayes are comparable to the centrally trained models. As such, there is generally no meaningfull difference in terms of AUC or AIC. Furthermore, the experiments show that it is possible to validate the model in a privacy-preserving manner despite it being impossible to efficiently classify an individual in a privacy-preserving manner.

It is however important to note that in certain edge-cases some validation approaches can show unreliable results. For example, the SVDG approach can cause problems when the test-folds are too small and the bins are generated on the fly while training, as opposed to working with pre-defined bins.If there are not enough individuals in the test-set to create multiple accurate bins on the fly this strategy will result in a loss of information. This was most notable when running the preliminary experiments with the Iris dataset, which is quite small.

Similarly, the model ran into problems using the SCV approach whenever the CPDs become too large because a node has multiple parents with a significant amount of bins. This is notable in the results of the Diabetes model. Certain nodes with multiple parents would end up with CPDs that contained more cells than there are individuals in the dataset. This led to an overestimation of the AUC if the SCV approach was used. However, this was not an issue when utilizing the SVDG approach, due to the stronger separation between training and validation data. Using larger bins can alleviate this problem to some extent. However, the bins cannot be made arbitrarily large as this will eventually cause a loss of information. Using expert-knowledge based discretization strategies tailored to each dataset, or a better automatic discretization strategy such as the MDL method mentioned earlier, would help avoid these problems.

These problems of overfitting and loss of information, show that it is extremely important to have an appropriate discretization strategy. So long as the potential pitfalls surrounding discretization are addressed, VertiBayes can be used to train and validate a Bayesian network in a vertical federated setting.

\subsection{Scalability}
Any algorithm that is adapted to a federated setting will be slower than the central counterpart due to the overhead caused by the protocols used to protect privacy. Our experiments confirmed the potential issues we brought up in section \ref{timesection} when we discussed the theoretical time complexity.

The effects of population-size proved to be negligible. The effects of the complexity of the network structure, that is to say the number of nodes and links within the network, is only relevant in so much that it creates more probabilities to calculate. As such, it runs in comparable time as the experiment that simply adds a few additional values to reach the same number of probabilities. As expected, the size of the CPDs that had to be calculated had the greatest effect on the total runtime.

This does mean that there are practical limitations to using VertiBayes, as it may take too long to train a large network. However, it should be noted that in certain settings a longer runtime might still be acceptable. For example, it is perfectly acceptable that training a model for use in a clinical setting takes an extended amount of time.

\subsection{Sensitive information in published Bayesian networks}
Publishing a Bayesian network, or any machine learning model, will reveal certain information about the training data, regardless of how the network is trained. When publishing a Bayesian network two important aspects will be revealed: the network structure and the CPDs. The network structure will only reveal which conditional dependencies exist amongst attributes, which is not sensitive data in most scenarios. The CPDs on the other hand, can potentially be used to reconstruct individual level data from the trainingset, when the probabilities in the CPD’s are based on one, or a few individuals. An effective countermeasure is using k-anonymity\cite{sweeney_k-anonymity_2002} to ensure that each probability in the CPDs represents a minimum amount of samples and that no probabilities of $0$ or $1$ are present in the CPDs. Such probabilities make it considerably easier to deduce individual level data, and they can also lead to artifacts when using the network. Lastly, a public Bayesian network can be used by one of the parties that participated in the training to guess (although not reconstruct) the data of the other parties based on their own data. Similarly, any third party with partial data can use the final model to estimate the missing values in his dataset. This is unavoidable and it should be taken into consideration when decisions are made about which models are to be made public. 

These concerns imply that there are practical limits to what privacy preserving techniques should  aim for. Trying to prevent any and all privacy issues using privacy preserving techniques during the training phase is futile when models are made public as the models themselves will always reveal some information.

\section{Conclusion}
In this paper, we have proposed a novel approach to train Bayesian network parameters from a vertically partitioned data with and without missing values. This method can deal with an arbitrary number of parties, only limited by the runtime. We have shown that there are no additional privacy risks compared to a centrally trained model beyond the ones presented by the specific privacy preserving scalar product protocol implementation used. Our experiments show there are no meaningful differences in performance between models trained with VertiBayes and models trained centrally, provided continuous variables are adequately discretized. They also show it is possible to estimate the performance of a model with vertically partitioned data with a reasonable accuracy. As such, VertiBayes is a useful tool for training Bayesian networks in a vertically partitioned setting.

\section{Future Work}
When using the model in a federated setting with a vertical split, it is currently not possible to efficiently classify or predict a new instance in a privacy preserving manner using a Bayesian network. It would be beneficial if a solution for this was found and implemented. In addition to this, VertiBayes could be improved by implementing more advanced discretization methods, such as MDL, in a vertically partitioned setting as our current implementation relies either on a very basic automatic discretization approach or the use of experts, which may not be the best discretization approach possible. Lastly, the impact of different missing data mechanisms on our proposed approach should be investigated. In this article we used data that missed completely at random, however, we did not look at other missing mechanisms, such as missing at random. It would be worthwhile to investigate whether this significantly influences the performance of the resulting model.

\bibliographystyle{IEEETran}
\bibliography{VertiBayes}

\begin{thebibliography}{10}
\providecommand{\url}[1]{#1}
\csname url@samestyle\endcsname
\providecommand{\newblock}{\relax}
\providecommand{\bibinfo}[2]{#2}
\providecommand{\BIBentrySTDinterwordspacing}{\spaceskip=0pt\relax}
\providecommand{\BIBentryALTinterwordstretchfactor}{4}
\providecommand{\BIBentryALTinterwordspacing}{\spaceskip=\fontdimen2\font plus
\BIBentryALTinterwordstretchfactor\fontdimen3\font minus
  \fontdimen4\font\relax}
\providecommand{\BIBforeignlanguage}[2]{{%
\expandafter\ifx\csname l@#1\endcsname\relax
\typeout{** WARNING: IEEEtran.bst: No hyphenation pattern has been}%
\typeout{** loaded for the language `#1'. Using the pattern for}%
\typeout{** the default language instead.}%
\else
\language=\csname l@#1\endcsname
\fi
#2}}
\providecommand{\BIBdecl}{\relax}
\BIBdecl

\bibitem{li_review_2020}
\BIBentryALTinterwordspacing
L.~Li, Y.~Fan, M.~Tse, and K.-Y. Lin, ``\BIBforeignlanguage{en}{A review of
  applications in federated learning},''
  \emph{\BIBforeignlanguage{en}{Computers \& Industrial Engineering}}, vol.
  149, p. 106854, Nov. 2020. [Online]. Available:
  \url{https://www.sciencedirect.com/science/article/pii/S0360835220305532}
\BIBentrySTDinterwordspacing

\bibitem{kairouz_advances_2021}
\BIBentryALTinterwordspacing
P.~Kairouz, H.~B. McMahan, B.~Avent, A.~Bellet, M.~Bennis, A.~N. Bhagoji,
  K.~Bonawitz, Z.~Charles, G.~Cormode, R.~Cummings, R.~G.~L. D'Oliveira, S.~E.
  Rouayheb, D.~Evans, J.~Gardner, Z.~Garrett, A.~Gascón, B.~Ghazi, P.~B.
  Gibbons, M.~Gruteser, Z.~Harchaoui, C.~He, L.~He, Z.~Huo, B.~Hutchinson,
  J.~Hsu, M.~Jaggi, T.~Javidi, G.~Joshi, M.~Khodak, J.~Konečný, A.~Korolova,
  F.~Koushanfar, S.~Koyejo, T.~Lepoint, Y.~Liu, P.~Mittal, M.~Mohri, R.~Nock,
  A.~Özgür, R.~Pagh, M.~Raykova, H.~Qi, D.~Ramage, R.~Raskar, D.~Song,
  W.~Song, S.~U. Stich, Z.~Sun, A.~T. Suresh, F.~Tramèr, P.~Vepakomma,
  J.~Wang, L.~Xiong, Z.~Xu, Q.~Yang, F.~X. Yu, H.~Yu, and S.~Zhao,
  ``\BIBforeignlanguage{en}{Advances and {Open} {Problems} in {Federated}
  {Learning}},'' \emph{\BIBforeignlanguage{en}{arXiv:1912.04977 [cs, stat]}},
  Mar. 2021, arXiv: 1912.04977. [Online]. Available:
  \url{http://arxiv.org/abs/1912.04977}
\BIBentrySTDinterwordspacing

\bibitem{pearl_probabilistic_2014}
J.~Pearl, \emph{\BIBforeignlanguage{en}{Probabilistic {Reasoning} in
  {Intelligent} {Systems}: {Networks} of {Plausible} {Inference}}}.\hskip 1em
  plus 0.5em minus 0.4em\relax Elsevier, Jun. 2014, google-Books-ID:
  mn2jBQAAQBAJ.

\bibitem{yang_privacy-preserving_2006}
\BIBentryALTinterwordspacing
Z.~Yang and R.~Wright, ``\BIBforeignlanguage{en}{Privacy-{Preserving}
  {Computation} of {Bayesian} {Networks} on {Vertically} {Partitioned}
  {Data}},'' \emph{\BIBforeignlanguage{en}{IEEE Transactions on Knowledge and
  Data Engineering}}, vol.~18, no.~9, pp. 1253--1264, Sep. 2006. [Online].
  Available: \url{http://ieeexplore.ieee.org/document/1661515/}
\BIBentrySTDinterwordspacing

\bibitem{wright_privacy-preserving_2004}
\BIBentryALTinterwordspacing
R.~Wright and Z.~Yang, ``Privacy-preserving {Bayesian} network structure
  computation on distributed heterogeneous data,'' in \emph{Proceedings of the
  tenth {ACM} {SIGKDD} international conference on {Knowledge} discovery and
  data mining}, ser. {KDD} '04.\hskip 1em plus 0.5em minus 0.4em\relax New
  York, NY, USA: Association for Computing Machinery, Aug. 2004, pp. 713--718.
  [Online]. Available: \url{https://doi.org/10.1145/1014052.1014145}
\BIBentrySTDinterwordspacing

\bibitem{zhiqiang_yang_improved_2005}
\BIBentryALTinterwordspacing
{Zhiqiang Yang} and R.~Wright, ``Improved {Privacy}-{Preserving} {Bayesian}
  {Network} {Parameter} {Learning} on {Vertically} {Partitioned} {Data},'' in
  \emph{21st {International} {Conference} on {Data} {Engineering} {Workshops}
  ({ICDEW}'05)}.\hskip 1em plus 0.5em minus 0.4em\relax Tokyo, Japan: IEEE,
  2005, pp. 1196--1196. [Online]. Available:
  \url{http://ieeexplore.ieee.org/document/1647809/}
\BIBentrySTDinterwordspacing

\bibitem{ng_towards_2021}
\BIBentryALTinterwordspacing
I.~Ng and K.~Zhang, ``\BIBforeignlanguage{en}{Towards {Federated} {Bayesian}
  {Network} {Structure} {Learning} with {Continuous} {Optimization}},''
  \emph{\BIBforeignlanguage{en}{arXiv:2110.09356 [cs, stat]}}, Oct. 2021,
  arXiv: 2110.09356. [Online]. Available: \url{http://arxiv.org/abs/2110.09356}
\BIBentrySTDinterwordspacing

\bibitem{cooper_bayesian_1992}
\BIBentryALTinterwordspacing
G.~F. Cooper and E.~Herskovits, ``\BIBforeignlanguage{en}{A {Bayesian} method
  for the induction of probabilistic networks from data},''
  \emph{\BIBforeignlanguage{en}{Machine Learning}}, vol.~9, no.~4, pp.
  309--347, Oct. 1992. [Online]. Available:
  \url{https://doi.org/10.1007/BF00994110}
\BIBentrySTDinterwordspacing

\bibitem{koller_probabilistic_2009}
D.~Koller and N.~Friedman, \emph{\BIBforeignlanguage{en}{Probabilistic
  {Graphical} {Models}: {Principles} and {Techniques}}}.\hskip 1em plus 0.5em
  minus 0.4em\relax MIT Press, Jul. 2009, google-Books-ID: 7dzpHCHzNQ4C.

\bibitem{dempster_maximum_1977}
\BIBentryALTinterwordspacing
A.~P. Dempster, N.~M. Laird, and D.~B. Rubin, ``Maximum {Likelihood} from
  {Incomplete} {Data} via the {EM} {Algorithm},'' \emph{Journal of the Royal
  Statistical Society. Series B (Methodological)}, vol.~39, no.~1, pp. 1--38,
  1977, publisher: [Royal Statistical Society, Wiley]. [Online]. Available:
  \url{https://www.jstor.org/stable/2984875}
\BIBentrySTDinterwordspacing

\bibitem{lauritzen_em_1995}
\BIBentryALTinterwordspacing
S.~L. Lauritzen, ``\BIBforeignlanguage{en}{The {EM} algorithm for graphical
  association models with missing data},''
  \emph{\BIBforeignlanguage{en}{Computational Statistics \& Data Analysis}},
  vol.~19, no.~2, pp. 191--201, Feb. 1995. [Online]. Available:
  \url{https://www.sciencedirect.com/science/article/pii/0167947393E0056A}
\BIBentrySTDinterwordspacing

\bibitem{du_building_2002}
W.~Du and Z.~Zhan, ``Building decision tree classifier on private data,'' in
  \emph{Proceedings of the {IEEE} international conference on {Privacy},
  security and data mining - {Volume} 14}, ser. {CRPIT} '14.\hskip 1em plus
  0.5em minus 0.4em\relax AUS: Australian Computer Society, Inc., Dec. 2002,
  pp. 1--8.

\bibitem{du_privacy-preserving_2001}
\BIBentryALTinterwordspacing
W.~Du and M.~Atallah, ``\BIBforeignlanguage{en}{Privacy-preserving cooperative
  statistical analysis},'' in \emph{\BIBforeignlanguage{en}{Seventeenth
  {Annual} {Computer} {Security} {Applications} {Conference}}}.\hskip 1em plus
  0.5em minus 0.4em\relax New Orleans, LA, USA: IEEE Comput. Soc, 2001, pp.
  102--110. [Online]. Available:
  \url{http://ieeexplore.ieee.org/document/991526/}
\BIBentrySTDinterwordspacing

\bibitem{goos_secure_2001}
\BIBentryALTinterwordspacing
M.~J. Atallah and W.~Du, ``Secure {Multi}-party {Computational} {Geometry},''
  in \emph{Algorithms and {Data} {Structures}}, G.~Goos, J.~Hartmanis, J.~van
  Leeuwen, F.~Dehne, J.-R. Sack, and R.~Tamassia, Eds.\hskip 1em plus 0.5em
  minus 0.4em\relax Berlin, Heidelberg: Springer Berlin Heidelberg, 2001, vol.
  2125, pp. 165--179, series Title: Lecture Notes in Computer Science.
  [Online]. Available: \url{http://link.springer.com/10.1007/3-540-44634-6_16}
\BIBentrySTDinterwordspacing

\bibitem{hutchison_private_2005}
\BIBentryALTinterwordspacing
B.~Goethals, S.~Laur, H.~Lipmaa, and T.~Mielikäinen,
  ``\BIBforeignlanguage{en}{On {Private} {Scalar} {Product} {Computation} for
  {Privacy}-{Preserving} {Data} {Mining}},'' in
  \emph{\BIBforeignlanguage{en}{Information {Security} and {Cryptology} –
  {ICISC} 2004}}, D.~Hutchison, T.~Kanade, J.~Kittler, J.~M. Kleinberg,
  F.~Mattern, J.~C. Mitchell, M.~Naor, O.~Nierstrasz, C.~Pandu~Rangan,
  B.~Steffen, M.~Sudan, D.~Terzopoulos, D.~Tygar, M.~Y. Vardi, G.~Weikum, C.-s.
  Park, and S.~Chee, Eds.\hskip 1em plus 0.5em minus 0.4em\relax Berlin,
  Heidelberg: Springer Berlin Heidelberg, 2005, vol. 3506, pp. 104--120, series
  Title: Lecture Notes in Computer Science. [Online]. Available:
  \url{http://link.springer.com/10.1007/11496618_9}
\BIBentrySTDinterwordspacing

\bibitem{vaidya_privacy_2002}
\BIBentryALTinterwordspacing
J.~Vaidya and C.~Clifton, ``Privacy preserving association rule mining in
  vertically partitioned data,'' in \emph{Proceedings of the eighth {ACM}
  {SIGKDD} international conference on {Knowledge} discovery and data mining},
  ser. {KDD} '02.\hskip 1em plus 0.5em minus 0.4em\relax New York, NY, USA:
  Association for Computing Machinery, Jul. 2002, pp. 639--644. [Online].
  Available: \url{https://doi.org/10.1145/775047.775142}
\BIBentrySTDinterwordspacing

\bibitem{van_daalen_privacy_2022}
\BIBentryALTinterwordspacing
F.~van Daalen, I.~Bermejo, L.~Ippel, and A.~Dekkers, ``Privacy preserving
  n-party scalar product protocol.'' [Online]. Available:
  \url{http://arxiv.org/abs/2112.09436}
\BIBentrySTDinterwordspacing

\bibitem{berlingerio_privacy_2019}
\BIBentryALTinterwordspacing
N.~C. Abay, Y.~Zhou, M.~Kantarcioglu, B.~Thuraisingham, and L.~Sweeney,
  ``Privacy preserving synthetic data release using deep learning,'' in
  \emph{Machine Learning and Knowledge Discovery in Databases}, M.~Berlingerio,
  F.~Bonchi, T.~Gärtner, N.~Hurley, and G.~Ifrim, Eds.\hskip 1em plus 0.5em
  minus 0.4em\relax Springer International Publishing, vol. 11051, pp.
  510--526, series Title: Lecture Notes in Computer Science. [Online].
  Available: \url{http://link.springer.com/10.1007/978-3-030-10925-7_31}
\BIBentrySTDinterwordspacing

\bibitem{moncada-torres_vantage6_2020}
A.~Moncada-Torres, F.~Martin, M.~Sieswerda, J.~Van~Soest, and G.~Geleijnse,
  ``\BIBforeignlanguage{eng}{{VANTAGE6}: an open source {priVAcy} {preserviNg}
  {federaTed} {leArninG} {infrastructurE} for {Secure} {Insight} {eXchange}},''
  \emph{\BIBforeignlanguage{eng}{AMIA ... Annual Symposium proceedings. AMIA
  Symposium}}, vol. 2020, pp. 870--877, 2020.

\bibitem{frank_data_2016}
E.~Frank, I.~H. Witten, and M.~A. Hall, \emph{\BIBforeignlanguage{en}{Data
  {Mining}, {Fourth} {Edition}:{Practical} {Machine} {Learning} {Tools} and
  {Techniques} {\textbar} {Guide} books}}.\hskip 1em plus 0.5em minus
  0.4em\relax Morgan Kaufmann Publishers Inc, 2016.

\bibitem{de_marsico_mobile_2015}
\BIBentryALTinterwordspacing
M.~De~Marsico, M.~Nappi, D.~Riccio, and H.~Wechsler,
  ``\BIBforeignlanguage{en}{Mobile {Iris} {Challenge} {Evaluation}
  ({MICHE})-{I}, biometric iris dataset and protocols},''
  \emph{\BIBforeignlanguage{en}{Pattern Recognition Letters}}, vol.~57, pp.
  17--23, May 2015. [Online]. Available:
  \url{https://www.sciencedirect.com/science/article/pii/S0167865515000574}
\BIBentrySTDinterwordspacing

\bibitem{lauritzen_local_1988}
\BIBentryALTinterwordspacing
S.~L. Lauritzen and D.~J. Spiegelhalter, ``Local computations with
  probabilities on graphical structures and their application to expert
  systems,'' vol.~50, no.~2, pp. 157--194, \_eprint:
  https://onlinelibrary.wiley.com/doi/pdf/10.1111/j.2517-6161.1988.tb01721.x.
  [Online]. Available:
  \url{https://onlinelibrary.wiley.com/doi/abs/10.1111/j.2517-6161.1988.tb01721.x}
\BIBentrySTDinterwordspacing

\bibitem{beinlich_alarm_1989}
\BIBentryALTinterwordspacing
I.~A. Beinlich, H.~J. Suermondt, R.~M. Chavez, and G.~F. Cooper,
  ``\BIBforeignlanguage{en}{The {ALARM} {Monitoring} {System}: {A} {Case}
  {Study} with two {Probabilistic} {Inference} {Techniques} for {Belief}
  {Networks}},'' \emph{\BIBforeignlanguage{en}{AIME 89}}, pp. 247--256, 1989,
  publisher: Springer, Berlin, Heidelberg. [Online]. Available:
  \url{https://link.springer.com/chapter/10.1007/978-3-642-93437-7_28}
\BIBentrySTDinterwordspacing

\bibitem{smith_using_1988}
\BIBentryALTinterwordspacing
J.~W. Smith, J.~Everhart, W.~Dickson, W.~Knowler, and R.~Johannes, ``Using the
  {ADAP} {Learning} {Algorithm} to {Forecast} the {Onset} of {Diabetes}
  {Mellitus},'' \emph{Proceedings of the Annual Symposium on Computer
  Application in Medical Care}, pp. 261--265, Nov. 1988. [Online]. Available:
  \url{https://www.ncbi.nlm.nih.gov/pmc/articles/PMC2245318/}
\BIBentrySTDinterwordspacing

\bibitem{fayyad_multi-interval_1993}
U.~M. Fayyad and K.~B. Irani, ``Multi-{Interval} {Discretization} of
  {Continuous}-{Valued} {Attributes} for {Classification} {Learning},'' in
  \emph{Ijcai.}, 1993, pp. 1022--1029.

\bibitem{akaike_new_1974}
H.~Akaike, ``A new look at the statistical model identification,'' \emph{IEEE
  Transactions on Automatic Control}, vol.~19, no.~6, pp. 716--723, Dec. 1974,
  conference Name: IEEE Transactions on Automatic Control.

\bibitem{spirtes_algorithm_1991}
P.~Spirtes, C.~N. Glymour, P.~Spirtes, and C.~Glymour, ``An algorithm for fast
  recovery of sparse causal graphs,'' \emph{Social Science Computer Review},
  pp. 62--72, 1991.

\bibitem{sweeney_k-anonymity_2002}
\BIBentryALTinterwordspacing
L.~Sweeney, ``k-{ANONYMITY}: {A} {MODEL} {FOR} {PROTECTING} {PRIVACY},''
  \emph{International Journal of Uncertainty, Fuzziness and Knowledge-Based
  Systems}, vol.~10, no.~05, pp. 557--570, Oct. 2002, publisher: World
  Scientific Publishing Co. [Online]. Available:
  \url{https://www.worldscientific.com/doi/abs/10.1142/S0218488502001648}
\BIBentrySTDinterwordspacing

\end{thebibliography}

\begin{IEEEbiography}[{\includegraphics[width=1in,height=1.25in,clip,keepaspectratio]{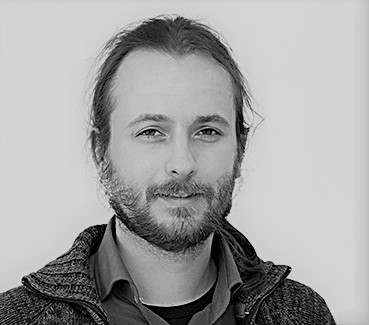}}]{Florian van Daalen}
	Florian van Daalen received his BSc degree in Knowledge Engineering from University Maastricht in 2012 and his MSc degree in Artificial Intelligence in 2014. He is currently working toward the PhD degree in Clinical Data Science within the Clinical Data Science group, University Maastricht, Netherlands. His research interests include privacy preserving techniques, federated learning, and ensemble based learning.
\end{IEEEbiography}

\begin{IEEEbiography}[{\includegraphics[width=1in,height=1.25in,clip,keepaspectratio]{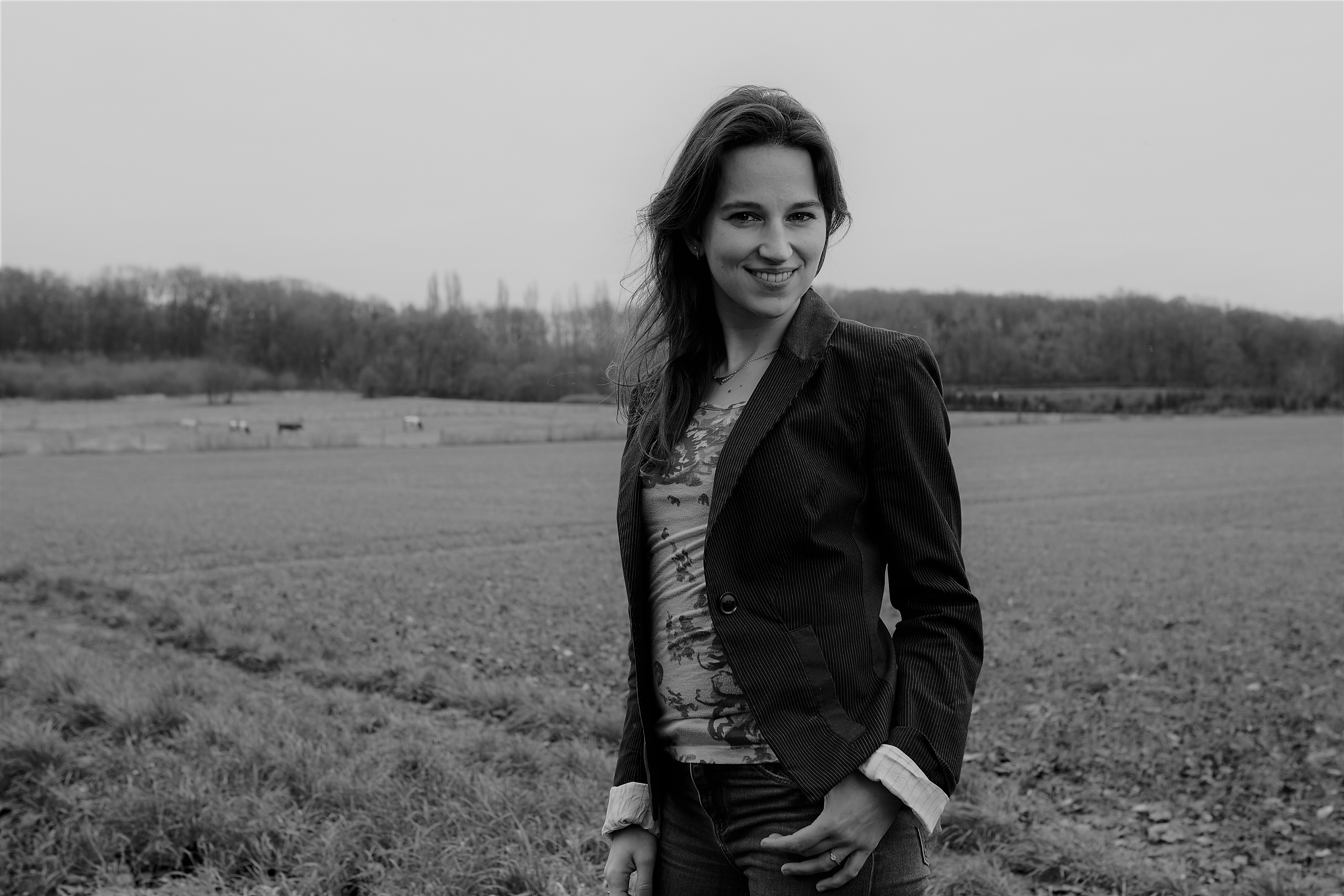}}]{Lianne Ippel}
	Lianne Ippel received her PhD in Statistics from Tilburg University on analyzing data streams with dependent observations, for which she won the dissertation award from General Online Research conference (2018). After a Postdoc at Maastricht University, she now works at Statistics Netherlands where she works at the methodology department on international collaborations and innovative methods for primary data collection.
\end{IEEEbiography}

\begin{IEEEbiography}[{\includegraphics[width=1in,height=1.25in,clip,keepaspectratio]{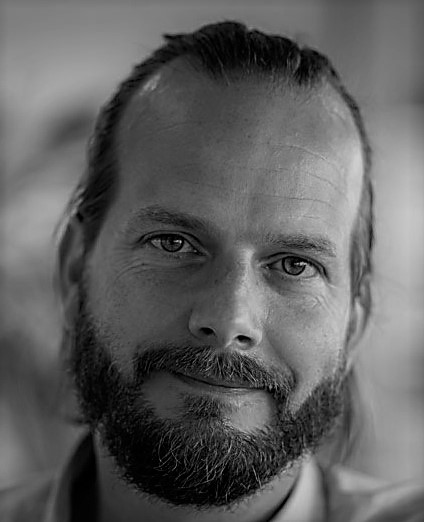}}]{Andre Dekker}
	Prof. Andre Dekker, PhD (1974) is a medical physicist and professor of Clinical Data Science at Maastricht University Medical Center and Maastro Clinic in The Netherlands. His Clinical Data Science research group (50 staff) focuses on 1) federated FAIR data infrastructures, 2) AI for health outcome prediction models and 3) applying AI to improve health. Prof. Dekker has authored over 200 publications, mentored more than 30 PhD students and holds multiple awards and patents on the topic of federated data and AI. He has held visiting scientist appointments at universities and companies in the UK, Australia, Italy, USA and Canada.
\end{IEEEbiography}

\begin{IEEEbiography}[{\includegraphics[width=1in,height=1.25in,clip,keepaspectratio]{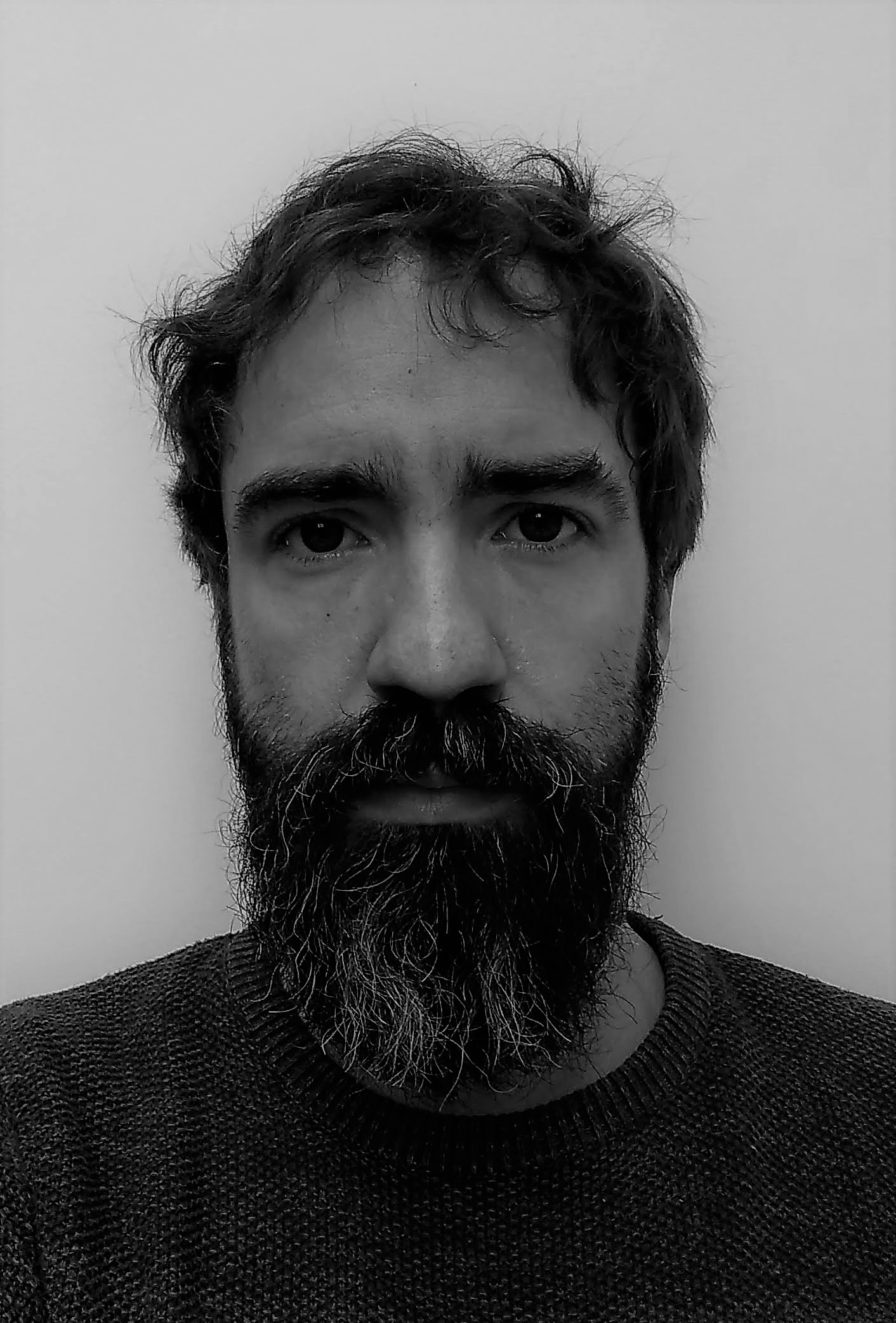}}]{Inigo Bermejo}
	Inigo Bermejo received the BSc degree on Computer Engineering from the University of the Basque Country, Spain, in 2006 and the PhD in Intelligent Systems from UNED, Spain, in 2015. He is currently a postdoctoral researcher at the Clinical Data Science group, Maastricht University. His research interests include privacy preserving techniques, prediction modelling and causal inference. 
\end{IEEEbiography}

\end{document}